\documentclass[letterpaper, 10 pt, conference]{ieeeconf}  

\IEEEoverridecommandlockouts                              
\usepackage[para,online,flushleft]{threeparttable}
\usepackage{algorithm}
\usepackage[noend]{algorithmic}
\usepackage{booktabs}
\usepackage{multirow}
\usepackage{adjustbox}
\usepackage{graphicx}
\usepackage{siunitx}
\usepackage[utf8]{inputenc}
\usepackage[T1]{fontenc}
\usepackage{amsfonts}
\usepackage{gensymb}

\usepackage{amsthm}

\usepackage{epsfig} 
\usepackage{amsmath} 
\usepackage{amssymb}  
\usepackage{siunitx}
\usepackage{subcaption}
\usepackage{caption}
\usepackage{color}
\usepackage[table]{xcolor}
\usepackage{soul}
\usepackage{theoremref}
\usepackage[noadjust]{cite}
\usepackage{amsthm}
\usepackage{mathtools}
\DeclarePairedDelimiter{\ceil}{\lceil}{\rceil}

\usepackage{hyperref}
\hypersetup{
    colorlinks=true,
    linkcolor=blue,
    filecolor=magenta,      
    urlcolor=blue,
    pdftitle={Overleaf Example},
    pdfpagemode=FullScreen,
    }

\newtheorem*{theorem*}{Theorem}

\title{\LARGE\bf ReDUCE: Reformulation of Mixed Integer Programs using Data from Unsupervised Clusters for Learning Efficient Strategies}
\author{Xuan Lin$^{1}$, Gabriel I.~Fernandez$^{1}$, and Dennis W.~Hong$^{1}$
\thanks{$^{1}$X. Lin, Gabriel I.~Fernandez, and Dennis W.~Hong are with the Robotics and Mechanisms Laboratory, Department of Mechanical and Aerospace Engineering, University of California, Los Angeles, CA 90095, USA.
        {\tt\small \{maynight,gabriel808,dennishong\}@ucla.edu}}
}

\begin{document}
\maketitle
\thispagestyle{empty}
\pagestyle{empty}

\begin{abstract}
Mixed integer convex and nonlinear programs, MICP and MINLP, are expressive but require long solving times. Recent work that combines learning methods on solver heuristics has shown potential to overcome this issue allowing for applications on larger scale practical problems. Gathering sufficient training data to employ these methods still present a challenge since getting data from traditional solvers are slow and newer learning approaches still require large amounts of data. In order to scale up and make these hybrid learning approaches more manageable we propose ReDUCE, a method that exploits structure within small to medium size datasets. We also introduce the bookshelf organization problem as an MINLP as a way to measure performance of solvers with ReDUCE. Results show that existing algorithms with ReDUCE can solve this problem within a few seconds, a significant improvement over the original formulation. ReDUCE is demonstrated as a high level planner for a robotic arm for the bookshelf problem.

\end{abstract}
%
%

\section{Introduction}
\label{Sec:introduction}
Optimization-based methods are useful tools for solving robotic motion planning problems. Typical approaches such as mixed-integer convex programs (MICPs) \cite{deits2014footstep,lin2019optimization}, nonlinear or nonconvex programs (NLPs) \cite{dai2014whole,winkler2018gait,shirai2020risk} and mixed-integer NLPs (MINLPs) \cite{soler2011route} offer powerful tools to formulate these problems. However, each has its own drawbacks. NLPs tend to suffer from local optimal solutions. In practice, local optimal solutions can sometimes have bad properties, such as inconsistent behavior as they depend on initial guesses.  Mixed-integer programs (MIPs) are a type of NP-hard problem. Branch-and-bound is usually used to solve MIPs \cite{boyd2007branch}. MIP solvers seek global optimal solutions, therefore, having more consistent behavior than NLP solvers. For small-scale problems, these algorithms usually find optimal solutions within a reasonable time \cite{lin2019optimization,tordesillas2019faster}. On the contrary, MIPs can require impractically long solving times for problems with a large number of integer variables \cite{lin2021designing}. MINLPs incorporate both integer variables and nonlinear constraints, hence, very expressive. Unfortunately, we lack efficient algorithms to tackle MINLPs. Many practical problems require a solving speed of at most a few seconds. As a result, it is difficult to implement most of the optimization schemes online for larger-scale problems.

\begin{figure}[!t]
		\centering
		\hspace*{-0.45cm}
		\includegraphics[scale=0.28]{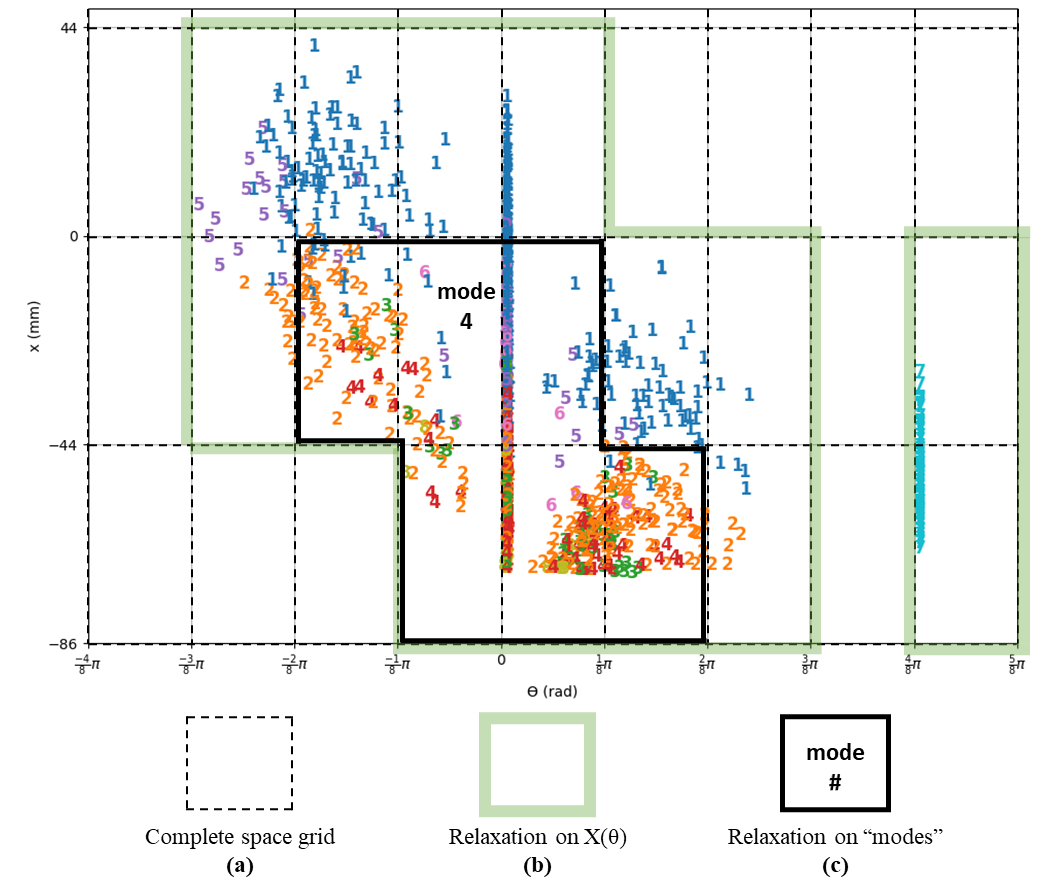}
		\caption {Angle and x position of solutions to the bookshelf problem's furthest left book are depicted by colored numbers indicating cluster membership. Each box on the grid corresponds to a different relaxation scheme using integer variables: (a) full relaxation of the whole space, (b) relaxation over the space with data, (c) relaxation for specific modes from clusters. For visualization purposes only mode 4 region is explicitly displayed. Notice how different modes can reduce the amount of integer variables needed.}
		\label{Fig:Figure_1}
\end{figure}

Recently, researchers have started to investigate machine learning methods to gather problem specific heuristics and speed up the MIP solving process. Standard algorithms to solve MIPs such as branch-and-bound and cutting plane methods rely on heuristics to quickly remove infeasible regions. Learning methods can be used to acquire better heuristics. For example, \cite{nair2020solving} used graph neural networks to learn heuristics. \cite{tang2020reinforcement} used reinforcement learning to discover efficient cutting planes. On the other hand, data can be collected to learn and solve specific problems \cite{zhu2019fast,cauligi2021coco}. In \cite{cauligi2021coco}, the authors proposed CoCo which collects problem features and solved integer variable strategies offline and trains a neural network. Effectively, this becomes a classification problem, where each strategy has a unique label. For online solving, the neural network will propose candidate solutions, reducing it to a convex program. The results show that CoCo can solve MIPs with around 50 integer variables from within a second. However, with more integer variables, CoCo suffers in two aspects. First, the number of possible strategies equals to $2^\lambda$, where $\lambda$ is the number of integer variables. As a result the number of unique strategies tend to be close to the total amount of data. This induces overfitting. Second, as the amount of integer variables increase, the solving speed dramatically slows. As a consequence, it is difficult to collect enough reliable training data.

In this paper, we propose ReDUCE, an algorithm that combines previous unsupervised learning work \cite{lin2021designing} with supervised learning, e.g., CoCo \cite{cauligi2021coco}, to solve larger-scale MICPs and MINLPs online. Unsupervised learning is employed on a small amount of initial data to retrieve sub-regions, clusters, inside the solution space. Integer variables are then assigned to each cluster. This allows us to retrieve important regions on the solution manifold and reduce the amount of integer variables needed. This then allows for fast generation of much larger datasets to train supervised learners on. All datasets generated can then be used to train a final learning model which allows interpolation between clusters. We also introduce the bookshelf organization problem in this paper to demonstrate ReDUCE's capabilities. Given a bookshelf with several books on top, an additional book needs to be placed on the shelf with minimal disturbance on the existing books. The bookshelf problem works well as a good benchmark because it: 1) is an MINLP that can be converted to an MICP problem with hundreds of integer variables, 2) can easily be scaled to push algorithms to their limit, and 3) has practical significance where data can reasonably be gathered, such as in the logistics industry.

To summarize, our contributions are as follows:
\begin{enumerate}
    \item Extend supervised learning schemes, e.g., CoCo, to solve MINLPs, where nonlinear or non-convex constraints are converted into MICPs constraints using convex envelopes,
    \item Use unsupervised learning to formulate the mixed-integer envelope constraints, significantly speeding up the data collection process for problems unsuitable for MICPs, i.e., very slow solving speeds, and
    \item Formulate the bookshelf organization problem as an MINLP and solve it within seconds with ReDUCE. 
\end{enumerate}


\section{Bookshelf Organization Problem Setup}
\label{Sec:problem_setup}
\begin{figure*}[!t]
		\centering
		\hspace*{-0.25cm}
		\includegraphics[scale=0.52]{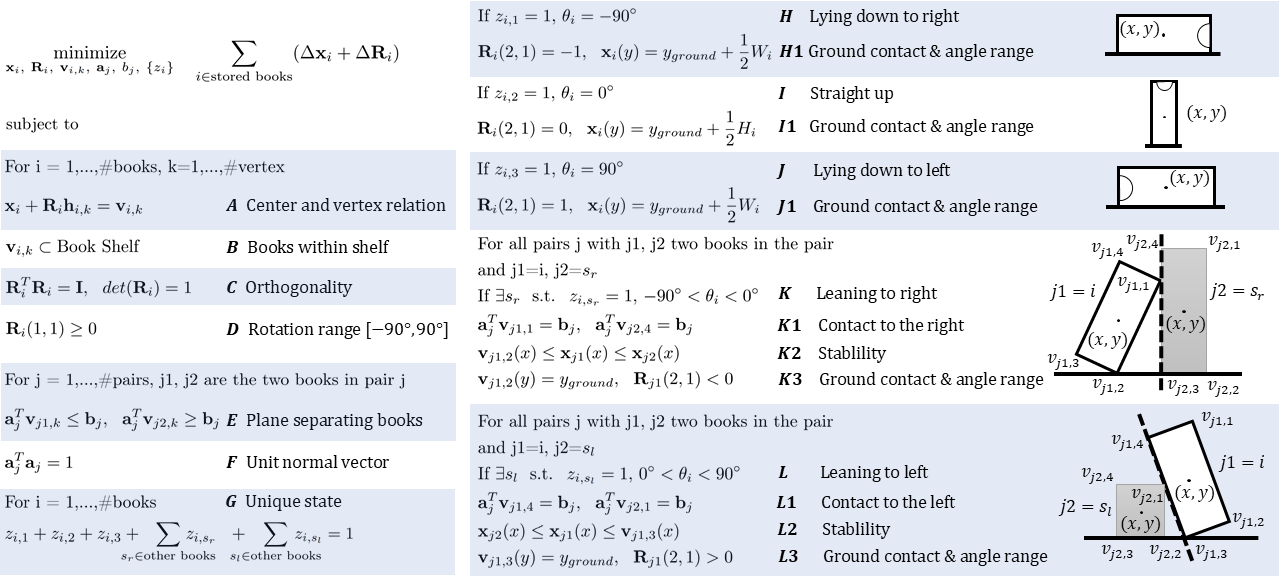}
		\caption {Complete formulation of the bookshelf organization problem.}
		\label{Fig:Complete_formulation}
\end{figure*}

Assume a 2D bookshelf with limited width $W$ and height $H$ contains rectangular books where book $i$ has width $W_{i}$ and height $H_{i}$ for $i=1,...,N-1$. A new book, $i=N$, is to be inserted into the shelf. The bookshelf contains enough books in various orientations, where in order to insert book $N$, the other $N-1$ books may need to be moved, i.e., optimize for minimal movement of $N-1$ books. This problem is useful in the logistics industry, such as robots filling shelves.



Fig. \ref{Fig:Complete_formulation} shows the constraints, variables and objective function for the bookshelf problem. The variables that characterize book $i$ are: position $\textbf{x}_{i} = [x_{i}, y_{i}]$ and angle $\theta_{i}$ about its centroid. $\theta_{i}=0$ when a book stands upright. The rotation matrix is: $\textbf{R}_{i} = [cos(\theta_{i}), \ -sin(\theta_{i}); \ sin(\theta_{i}), \ cos(\theta_{i})]$. Let the 4 vertices of book $i$ be $\textbf{v}_{i,k}$, $k=1,2,3,4$. The constraint \textbf{A} in Fig. \ref{Fig:Complete_formulation} shows the linear relationship between $\textbf{x}_{i}$ and $\textbf{v}_{i,k}$, where $\textbf{h}_{i,k}$ is the constant offset vector from its centroid to vertices. Constraint \textbf{B} enforces that all vertices of all books stay within the bookshelf, a linear constraint. Constraint \textbf{C} enforces the orthogonality of the rotation matrix, a bilinear (non-convex) constraint. Constraint \textbf{D} enforces that the angle $\theta_{i}$ stays within $[-90\degree, 90\degree]$, storing books right side up.

To ensure that the final book positions and orientations do not overlap with each other, separating plane constraints are enforced. For convex shapes, the two shapes do not overlap with each other if and only if there exists a separating hyperplane $\textbf{a}^{T}\textbf{x}=b$ in between \cite{boyd2004convex}. That is, for any point $\textbf{p}_{1}$ inside shape 1 then $\textbf{a}^{T}\textbf{p}_{1} \leq b$, and for any point $\textbf{p}_{2}$ inside shape 2 then $\textbf{a}^{T}\textbf{p}_{2} \geq b$. This is represented by constraint \textbf{E}. Constraint \textbf{F} enforces $\textbf{a}$ to be a normal vector. Both \textbf{E} and \textbf{F} are bilinear constraints.



Finally, we need to assign a state to each book $i$. For each book, it can be standing straight up, laying down on its left or right, or leaning towards left or right against some other book, as shown in the far right column in Fig. \ref{Fig:Complete_formulation}. For each book $i$, we assign a set of integer variables ${z_{i}}$. If book $i$ stands upright ($z_{i,2}=1$) or lays flat on its left ($z_{i,1}=1$) or right ($z_{i,3}=1$), constraints \textbf{I1} or \textbf{J1} or \textbf{H1} are enforced, respectively. If book $i$ leans against another book on the left or right, constraints in \textbf{K} and \textbf{L} are enforced, respectively. To this end checks need to indicate the contact between books. By looking at the right column in Fig. \ref{Fig:Complete_formulation}, we can reasonably assume that the separating plane $\textbf{a}^{T}\textbf{x}=b$ always crosses vertex 1 of the book on the left and vertex 4 of the book on the right. This is represented by bilinear constraints, \textbf{K1} and \textbf{L1}. In addition, the books need to remain stable given gravity. Constraints \textbf{K2} and \textbf{L2} enforce that a book is stable if its $x$ position stays between the supporting point of itself (vertex 2 if leaning rightward and vertex 3 if leaning leftward) and the $x$ position of the book that it is leaning onto. Lastly, constraint \textbf{K3} and \textbf{L3} enforce that the books have contact with the \emph{ground}. For practical reasons, we assume that books cannot stack onto each other, i.e, each book has to touch the \emph{ground} of bookshelf at at least one point. We note that constraints in \textbf{H}, \textbf{I}, \textbf{J}, \textbf{K}, and \textbf{L} can be easily formulated as MICP constraints using big-M formulation \cite{vielma2015mixed}, such that they are enforced only if the associated integer variable $z=1$. Also, it can easily be extended to allow stacking for our problem. Any contact conditions between pairs of books may also be added into this problem as long as it can be formulated as mixed-integer convex constraint. Overall, this is a problem with integer variables, ${z_{i}}$, and non-convex constraints \textbf{C}, \textbf{E}, \textbf{F}, \textbf{K1}, and \textbf{L1}, hence, an MINLP problem.

\begin{figure}[!t]
		\centering
		\hspace*{-0.35cm}
		\includegraphics[scale=0.52]{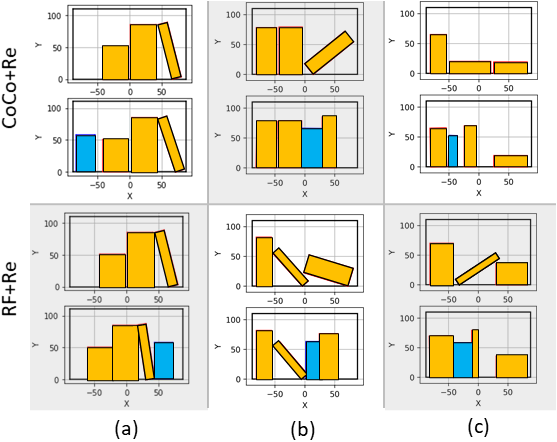}
		\caption {Solved scenes of bookshelves. \emph{Top Row}: Instance solved by CoCo with ReDUCE. \emph{Bottom Row}: Instance solved by RF with ReDUCE. Column (a), (b), (c) corresponds to cluster 0, 1, 2, respectively. Each cell contains a before and after scene with the upper diagram showing the original bookshelf with orange rectangles representing stored books and the lower diagram showing the solved bookshelf where the blue rectangle represents the inserted book. For column (a), we demonstrate a scene where RF gives a much worse result than CoCo such that multiple books are moved. However, CoCo and RF usually give similar results.}
		\label{Fig:Solved_cases}
\end{figure}

Practically, this problem presents challenges for retrieving high quality solutions. If robots were used to store books, the permissible solving time is several seconds, and less optimal solutions means longer realization times. For example, in Fig. \ref{Fig:Solved_cases} a non-optimal insertion induces multiple additional robot motions that dramatically increase the chance of failure. There are several potential approaches to resolving this issue: fix one set of nonlinear variables and solve MICP \cite{posa2015stability}, convert the nonlinear constraints into piece-wise linear constraints and formulate them into an MICP \cite{deits2014footstep}, or directly applied MINLP solvers such as BONMIN \cite{bonami2007bonmin}. As expected, these approaches struggle to meet the requirements. In this paper, we implement ReDUCE to satisfy them.

\section{Learning Algorithm} 
\label{Sec:learning_algorithm}
Previous work demonstrated the potential of learning mixed-integer strategies offline and then using the learned model to sample candidate solutions and solve convex programs online \cite{zhu2019fast,cauligi2021coco,bertsimas2021voice}. In \cite{lin2021designing}, the authors proposed an unsupervised learning method to identify important regions in the solution space. This approach effectively reformulated the whole MIP into multiple problems with a reduced number of integer variables. ReDUCE further builds upon this notion and combines those two approaches.



Assume that we are given a set of problems parametrized by $\Theta$ that is drawn from a distribution $D(\Theta)$. For each $\Theta$, we seek a solution to the optimization problem:


\begin{equation}
\begin{aligned}
& \underset{\textbf{x}, \ \textbf{z}}{\text{minimize}} \ f_{obj}(\textbf{x}, \textbf{z}; \Theta) \\
\text{s. t.} \ \ & f_{i}(\textbf{x}, \textbf{z}; \Theta) \leq 0, \ \ i = 1,...,m_{f} \\
& b_{j}(\textbf{x}, \textbf{z}; \Theta) \leq 0, \ \ j = 1,...,m_{b}
\label{Eqn:General_formulation}
\end{aligned}
\end{equation}

Where $\textbf{x}$ denotes continuous variables and $\textbf{z}$ binary variables with $z_{i} \in \{0, 1\}$ for $i=1,...,dim(\textbf{z})$. Constraints $f_{i}$ are mixed-integer convex, meaning if the binary variables $\textbf{z}$ are relaxed into continuous variables $\textbf{z} \in [0, 1]$, $f_{i}$ becomes convex. Constraints $b_{j}$ are mixed-integer bilinear, meaning that relaxing the binary variables gives bilinear constraints. Without loss of generality, $\textbf{x}$ and $\textbf{z}$ are assumed to be involved in each constraint. We omit equality constraints in \eqref{Eqn:General_formulation} as they can be turned into two inequality constraints from opposite directions. Similar to \cite{lin2021designing}, the solution set with respect to fixed $\textbf{z}$, $X(\Theta_{i}; \textbf{z})$, is defined to be the manifold containing all feasible $\textbf{x}$ given $\textbf{z}$. If $\textbf{z}$ is infeasible, $X(\Theta_{i}; \textbf{z}) = \emptyset$. The full solution set, $X(\Theta_{i})$, for the problem embedded inside the solution space, $\mathbb{S}\in\mathbb{R}^{dim(\textbf{x})}$, is defined by $\cup_{\textbf{z}} X(\Theta_{i}; \textbf{z})\; \forall\, \textbf{z}$.

In this paper, we take the approach to convert bilinear constraints into mixed-integer linear constraints by gridding the solution space $\mathbb{S}$ and approximating the constraints locally inside grids with McCormick envelopes similar to \cite{dai2019global}. A McCormick envelope relaxation of one bilinear constraint $w=xy$ \cite{castro2015tightening} is the best linear approximation defined over a pair of lower and upper bounds $[x^{L}, x^{U}]$ and $[y^{L}, y^{U}]$. Therefore, we first assign grids $G(\textbf{x})$ to $\mathbb{S}$. Let $\{g_{l}\}$, $l=1,...,L$ be one cell in the grid with upper and lower bounds $[\textbf{x}^{L}, \textbf{x}^{U}]$. We introduce additional integer variables $\textbf{n}$, where $n_{i} \in \{0, 1\}$. Each unique value of $\textbf{n}$ corresponds to one cell in the grid within which McCormick envelope relaxations are applied. The constraints $b_{j}(\textbf{x}, \textbf{z}; \Theta) \leq 0$, $j=1,...,m_{b}$ are converted into $L$ constraints: $E_{b_{j},l}(\textbf{x}, \textbf{z}, \textbf{n}; \Theta) \leq 0$ for $j = 1,...,m_{b}$ and $l = 1,...,L$, turning (\ref{Eqn:General_formulation}) into an MICP:




\begin{equation}
\begin{split}
& \underset{\textbf{x}, \ \textbf{z}, \ \textbf{n}}{\text{minimize}} \ f_{obj}(\textbf{x}, \textbf{z}, \textbf{n}; \Theta)\\
\text{s. t.} \qquad\; 
f_{i}(\textbf{x}, \textbf{z}; \Theta) \leq 0, & \ \ i = 1,...,m_{f} \\
E_{b_{j},l}(\textbf{x}, \textbf{z}, \textbf{n}; \Theta)& \leq 0, \ \ j = 1,...,m_{b}, \ \ l = 1,...,L \qquad
\label{Eqn:Updated_formulation}
\end{split}
\end{equation}


To speed up the solving speed for MICPs, \cite{bertsimas2021voice,cauligi2021coco,bertsimas2019online} define integer strategies to be tuples of $\mathcal{I}(\Theta_{i}) = (\textbf{z}^{*}, \textbf{n}^{*}, \mathcal{T}(\Theta_{i}))$, where $(\textbf{x}^{*}, \textbf{z}^{*}, \textbf{n}^{*})$ is an optimizer for problem \eqref{Eqn:Updated_formulation}, $\mathcal{T}(\Theta_{i}) = \{i \in {1,...,m_{f}}, \ l \in 1,...,L| f_{i}(\textbf{x}^{*}, \textbf{z}^{*}; \Theta) = 0, \ E_{b_{j},l}(\textbf{x}^{*}, \textbf{z}^{*}, \textbf{n}^{*}; \Theta)=0 \}$ is the set of active inequality constraints. Given an optimal integer strategy $\mathcal{I}(\Theta_{i})$, solution to (\ref{Eqn:Updated_formulation}) can be retrieved through solving a convex optimization problem. If this approach returns a small amount of integer variables, it may perform well. However, approximating nonlinear constraints with mixed-integer convex constraints usually gives large number of integer variables when high approximation accuracy is desired \cite{dai2019global}. As a result, the solving time grows fast. This makes it difficult to collect data to train supervised learners.


This paper uses clustering method on a relatively smaller amount of pre-solved data to identify important regions on $X(\Theta_{i})$, and implement McCormick envelope relaxations around those regions which significantly improves the MIP solving speed. The nominal dimension of the solution space is $dim(\textbf{x})$. However, due to the existence of constraints, the actual solution set of $X(\Theta)$ is of much lower dimension. Relaxing the complete space of $\textbf{x}$ is unnecessary as many grid cells are infeasible or non-optimal and the MIP solver should not need to spend time in exploring those regions. Furthermore, we almost never need the complete $X(\Theta)$ manifold in practice. Many robotic systems operate under different modes corresponding to regions on $X(\Theta)$ where the optimal solutions $\textbf{x}^{*}$ populate. With data unsupervised learning can identify those regions. Within each region, we only need a smaller amount of integer variables $\textbf{z}, \textbf{n}$. Consequently, the solving speed can be significantly improved.


The detailed steps of ReDUCE are shown in Algorithm \ref{alg:alg}. To begin, ReDUCE requires a relatively small amount of pre-solved data $(\Theta_{k}, \textbf{x}_{k})$, $k=1,...,K$ termed \textit{kick off} data throughout. This data may be collected by solving the non-reduced formulation \eqref{Eqn:Updated_formulation}. If \eqref{Eqn:Updated_formulation} represents a practical problem, the data may come from simulation or human demonstration on real hardware. We also need to pre-assign grids $G(\textbf{x})$ to the solution space $\mathbb{S}$. The size of grids depends on the approximation accuracy requirement for bilinear constraints. ReDUCE begins by performing unsupervised learning on the \textit{kick off data} to retrieve clusters that indicates regions on $X(\theta)$. Density-Based Spatial Clustering of Applications with Noise (DBSCAN) \cite{ester1996density} was used to cluster on $\textbf{x}$ which gives clusters ${\textbf{x}_{1}}, {\textbf{x}_{2}}, ..., {\textbf{x}_{C}}$, where $C$ is the number of clusters. We then trace $\Theta \rightarrow \textbf{x}$ map backwards to create clusters in $\Theta$ space, i.e. $(\{\Theta\}_{1}, \{\textbf{x}\}_{1}), ..., (\{\Theta\}_{C}, \{\textbf{x}\}_{C})$. Next, a supervised classifier is trained to classify $(\Theta, c)$. We use random forest which requires relatively smaller amounts of data to train than deep learning methods. Thus, the amount of \emph{kick off} data, $K$, can be relatively small.

The main difficulty to generate training data for \eqref{Eqn:Updated_formulation} is its slow solving speed due to large amount of integer variables. If $\mathbb{R}^{dim(\textbf{x})}$ is segmented into smaller regions, e.g. clusters, the required integer variables for each cluster can be reduced. This may be seen in Fig. \ref{Fig:Figure_1} which is an instance of 2-dimensional (dim) $X(\Theta)$ from a bookshelf experiment described in Sec.~\ref{Sec:experiment_setup}. In this paper, we use a $log_{2}N$ formulation which means $N$ grids will be represented by $\ceil{log_{2}N}$ integer variables. For example, $17 \sim 32$ grids can be represented by 5 integer variables. In Fig. \ref{Fig:Figure_1}, the complete space has 27 grids which can be represented with at least 5 integer variables. However, mode 4 occupies only 6 grids which is represented with 3 integer variables. One advantage of using DBSCAN is that it uses a threshold to decide the boundaries of each cluster and identify outliers. Since DBSCAN is based off of densities, it is able to predict outliers from the data. Outliers can be removed from training data wherein the classifier can be used to make a prediction. Based on that prediction, it is possible that some outliers may have membership to multiple clusters. Also outliers may indicate insufficient sample size near that particular region. This may guide practitioners where to collect more data.

With the trained classifier, we can classify a much greater amount of problems $\Theta \sim D(\Theta)$ and solve them quickly within each clustered region to collect more training data, indicated in line 4-14 of Algorithm \ref{alg:alg}. For all $\{\textbf{x}\}_{c}$ in cluster $c$, we find the grid cells that they occupy and re-assign integer variables $\textbf{z}_{c}$ and $\textbf{n}_{c}$ to each cluster (line 8-10). Eventually all data are put together for training, so the definition of integer variables needs to be consistent across clusters. Thus, we recover the original integer variables $\textbf{z}$ and $\textbf{n}$ (line 13). With $\Theta$, $\textbf{z}_{c}$ and $\textbf{n}_{c}$ for cluster $c$, problem \eqref{Eqn:Updated_formulation} can be reformulated and solved. The feature $\Theta$, solution $\textbf{x}^{*}$, recovered integer variables $\textbf{z}^{*}$ and $\textbf{n}^{*}$ are added to the dataset. Finally, the dataset is used to train a strategy learner, e.g. CoCo.

The idea of solving smaller sub-MIPs bears similarities with algorithms such as RENS \cite{berthold2014rens} and Neural Diving \cite{nair2020solving} where a subset of integer variables are fixed from linear program solutions or a learned model, while the others are solved. Our method, however, segments the problem through an unsupervised clustering approach with the goal of generating enough training data for supervised learning.

\begin{algorithm}[t] 
\small 
\algsetup{linenosize=\small}
\caption{$\operatorname{ReDUCE}$}
\label{alg:alg}
\begin{tabular}{@{}l@{\ }l}
    \textbf{Input} & \text{Kick off data}  $\{(\Theta_{k},\; \textbf{x}_{k})\},\; k=1,...,K$,\\
     &\text{Grid} $G(\textbf{x}) \;\text{with cells}\; \{g_{l}\},\; l=1,...,L$, \\
     &\text{DBSCAN clustering threshold } $\epsilon$\\
     &$d_{c}$ desired amount of samples on label $c,\; c=1,...,C$
\end{tabular}
\begin{algorithmic}[1]
\STATE DBSCAN$(\{\textbf{x}\}, \epsilon)\longrightarrow \{\textbf{x}\}_{1},...,\{\textbf{x}\}_{C} $ 
\STATE $\{\{\Theta\}_c : \textbf{x}\in (\Theta, \textbf{x}) \land \{\textbf{x}\}_{c}\;\forall\; c=1,...,C\}$
\STATE Train Random Forest: RF$(\{\Theta\})\rightarrow \{c\}$
\WHILE{Samples $|\{\Theta\}_c| < d_{c}$}
\STATE Sample $\Theta \sim D(\Theta)$, Classify RF$(\Theta)$ and Store $(\Theta, c)$
\ENDWHILE
\FOR{$c=1,...,C$}
\STATE Initialize grid cell list $\{g_{c}\}$ and dataset $S_{c}$
\FOR{each point in $\{\textbf{x}\}_{c}$}
\STATE Find corresponding grid $g_{l}$ and add $g_{l}$ to $\{g_{c}\}$
\ENDFOR
\STATE Assign integer variables $\textbf{z}_{c}, \textbf{n}_{c}$ to $\{g_{c}\}$
\STATE Formulate $\mathcal{P}$ from \eqref{Eqn:Updated_formulation} with $\{\Theta\}_{c}$, replacing $\textbf{z}$, $\textbf{n}$ by $\textbf{z}_{c}, \textbf{n}_{c}$
\IF{$\mathcal{P}$ has solution $\textbf{z}_{c}^{*}, \textbf{n}_{c}^{*}, \textbf{x}^{*}$}
\STATE Recover original $\textbf{z}^{*}$, $\textbf{n}^{*}$ from $\textbf{z}_{c}^{*}, \textbf{n}_{c}^{*}$
\STATE Add $\Theta$, $\textbf{z}^{*}$, $\textbf{n}^{*}$, $\textbf{x}^{*}$ to $S_{c}$
\ENDIF
\ENDFOR
\STATE Initialize Neural Network or desired model: $f(\cdot)$
\STATE Use $S_{c}$ to train: $f(\Theta)\rightarrow (\textbf{z}^{*},\textbf{n}^{*})$
\RETURN $f(\cdot)$
\end{algorithmic} 
\end{algorithm}

\section{Experiment} 
\subsection{Experiment Setup}
\label{Sec:experiment_setup}
We use ReDUCE to solve the bookshelf organization problem. We place 3 books inside the shelf where 1 additional book is to be inserted. Grids are assigned to the variables involved in the non-convex constraint \textbf{C}, \textbf{E}, \textbf{F}, \textbf{K1}, \textbf{L1}: $\textbf{R}_{i}(\theta_{i})$, $\textbf{a}_{j}$ and $\textbf{v}_{i,k}$. These variables span a 48 dimensional space. The rotation angles $\theta_{i}$, which includes $\textbf{R}_{i}$, are gridded at a $\frac{\pi}{8}$ interval. Elements in $\textbf{a}_{j}$ are gridded on 0.25 intervals. Elements in $\textbf{v}_{i,k}$ are gridded at intervals $\frac{1}{4}$ the shelf width $W$ and height $H$. Since books are not allowed to stack on each other, there is an order of books from left to right. We order the feature and solution vector according to this order. We use an MICP formulation that has $log_{2}N$ integer variables (explained in detail in the appendix) where $N$ is the number of grids. This results in 130 integer variables in total. The feature vector includes the centroid positions, angles, heights and widths of stored books and height and width of the book to be inserted. The feature dimension is 17.


The \emph{kick off} data was collected using a 2-dim simulated environment of books on a shelf. Initially, 4 randomly sized books are arbitrarily placed on the shelf, and then 1 is randomly removed and regarded as the book to be inserted. Contrary to the sequence, the initial state with 4 books represents one feasible (not necessarily optimal) solution to the problem of placing a book on a shelf with 3 existing books. Since this problem can be viewed as high-level planning for robotic systems, the simulated data is sufficient. For applications outside of the scope of this paper real world data may be preferable in this pipeline.



\subsection{Unsupervised Learning}
\label{Sec:unsupervised_learning}
\begin{table*}[!h]
    \centering
    \setlength{\tabcolsep}{5.5pt}
    \begin{tabular}{c|c|c|c|c|c}
                      &\textbf{ReDUCE} &\textbf{CoCo+Re} &\textbf{RF+Re} &\textbf{Baseline+Re} &\textbf{MIP+Re} \\

\begin{tabular}{@{}c@{}}
\textbf{Cluster}   \\
\cellcolor{blue!10}
\textbf{0}         \\
\cellcolor{blue!10}\\
\cellcolor{blue!10}\\
\textbf{$\;$1}     \\                                     
                   \\
                   \\
\cellcolor{blue!10}
\textbf{2}         \\
\cellcolor{blue!10}\\
\cellcolor{blue!10}\\
\textbf{$\;$3}     \\
                   \\
\end{tabular}

                &\begin{tabular}{@{}ccc@{}}
                 \textbf{N/total} &                \textbf{\%} &           \textbf{Int} \\
    \cellcolor{blue!10} 997/1000  &\cellcolor{blue!10} 99.7\%  &\cellcolor{blue!10}77   \\
    \cellcolor{blue!10} 2984/2999 &\cellcolor{blue!10} 99.4\%  &\cellcolor{blue!10}77   \\ 
    \cellcolor{blue!10} 4947/5000 &\cellcolor{blue!10} 98.9\%  &\cellcolor{blue!10}77   \\
                   $\;$ 698/699   &               $\,$ 99.9\%  &                   77   \\
                   $\;$ 1996/1999 &               $\,$ 99.8\%  &                   77   \\ 
                   $\;$ 5976/5999 &               $\,$ 99.6\%  &                   77   \\
    \cellcolor{blue!10}  599/600  &\cellcolor{blue!10} 99.8\%  &\cellcolor{blue!10}66   \\
    \cellcolor{blue!10} 2970/2999 &\cellcolor{blue!10} 99.0\%  &\cellcolor{blue!10}66   \\
    \cellcolor{blue!10} 5476/5599 &\cellcolor{blue!10} 97.8\%  &\cellcolor{blue!10}66   \\
                   $\;$ 339/399   &               $\,$ 85.0\%  &                   46   \\
                   $\;$ 681/900   &               $\,$ 75.6\%  &                   46   \\
                \end{tabular}
                
                &\begin{tabular}{@{}crrr@{}}
                \textbf{S\%}       &              \textbf{Det} &           \textbf{Avg}  &            \textbf{Max} \\
        \cellcolor{blue!10} 93.4\% &\cellcolor{blue!10} 210    &\cellcolor{blue!10} 1.39 &\cellcolor{blue!10} 7.39 \\
        \cellcolor{blue!10} 97.0\% &\cellcolor{blue!10} 144    &\cellcolor{blue!10} 1.25 &\cellcolor{blue!10} 7.35 \\
        \cellcolor{blue!10} 98.0\% &\cellcolor{blue!10} 119    &\cellcolor{blue!10} 1.05 &\cellcolor{blue!10} 7.23 \\
                            91.2\% &                    148    &                    1.42 &                    7.37 \\
                            96.2\% &                     93    &                    1.30 &                    7.58 \\
                            99.0\% &                     73    &                    1.24 &                    7.12 \\
        \cellcolor{blue!10} 91.0\% &\cellcolor{blue!10}  95    &\cellcolor{blue!10} 1.36 &\cellcolor{blue!10} 6.70 \\  
        \cellcolor{blue!10} 96.8\% &\cellcolor{blue!10}  60    &\cellcolor{blue!10} 1.13 &\cellcolor{blue!10} 7.37 \\  
        \cellcolor{blue!10} 98.4\% &\cellcolor{blue!10}  47    &\cellcolor{blue!10} 1.07 &\cellcolor{blue!10} 7.47 \\  
                            94.0\% &                     28    &                    1.18 &                    7.03 \\  
                            97.2\% &                     20    &                    0.95 &                    6.92 \\  
                \end{tabular}
                                       
                    &\begin{tabular}{@{}crrr@{}}
                    \textbf{S\%}       &              \textbf{Det} &            \textbf{Avg} &            \textbf{Max} \\
            \cellcolor{blue!10} 93.2\% &\cellcolor{blue!10} 348    &\cellcolor{blue!10} 1.42 &\cellcolor{blue!10} 7.58 \\
            \cellcolor{blue!10} 97.0\% &\cellcolor{blue!10} 253    &\cellcolor{blue!10} 1.31 &\cellcolor{blue!10} 7.46 \\
            \cellcolor{blue!10} 96.0\% &\cellcolor{blue!10} 194    &\cellcolor{blue!10} 1.17 &\cellcolor{blue!10} 6.75 \\
                                90.0\% &                    267    &                    1.35 &                    7.10 \\
                                92.6\% &                    204    &                    1.27 &                    7.21 \\
                                96.8\% &                    169    &                    1.12 &                    7.57 \\
            \cellcolor{blue!10} 90.6\% &\cellcolor{blue!10} 122    &\cellcolor{blue!10} 1.33 &\cellcolor{blue!10} 7.51 \\  
            \cellcolor{blue!10} 93.8\% &\cellcolor{blue!10}  88    &\cellcolor{blue!10} 1.00 &\cellcolor{blue!10} 7.42 \\  
            \cellcolor{blue!10} 95.2\% &\cellcolor{blue!10}  84    &\cellcolor{blue!10} 1.04 &\cellcolor{blue!10} 6.97 \\  
                                92.6\% &                     37    &                    0.94 &                    7.04 \\  
                                97.8\% &                     29    &                    0.73 &                    6.94 \\  
                    \end{tabular}
                        
                        &\begin{tabular}{@{}cr@{}}
                        \textbf{S\%}       &            \textbf{Det}   \\
                \cellcolor{blue!10} 65.4\% &\cellcolor{blue!10} 856    \\
                \cellcolor{blue!10} 63.7\% &\cellcolor{blue!10} 862    \\
                \cellcolor{blue!10} 65.5\% &\cellcolor{blue!10} 933    \\
                                    57.4\% &                    608    \\
                                    60.0\% &                    554    \\
                                    60.5\% &                    543    \\
                \cellcolor{blue!10} 59.5\% &\cellcolor{blue!10} 281    \\  
                \cellcolor{blue!10} 57.0\% &\cellcolor{blue!10} 293    \\  
                \cellcolor{blue!10} 57.7\% &\cellcolor{blue!10} 251    \\  
                                    63.2\% &                    162    \\  
                                    57.5\% &                    132    \\  
                        \end{tabular}
                            
                            &\begin{tabular}{@{}rr@{}}
                            \textbf{Avg}     &            \textbf{Max}  \\
                    \cellcolor{blue!10} 4.52 &\cellcolor{blue!10} 37.86 \\
                    \cellcolor{blue!10} 4.52 &\cellcolor{blue!10} 37.86 \\
                    \cellcolor{blue!10} 4.52 &\cellcolor{blue!10} 37.86 \\
                                        2.27 &                    16.61 \\
                                        2.27 &                    16.61 \\
                                        2.27 &                    16.61 \\
                    \cellcolor{blue!10} 0.92 &\cellcolor{blue!10} 5.46  \\  
                    \cellcolor{blue!10} 0.92 &\cellcolor{blue!10} 5.46  \\  
                    \cellcolor{blue!10} 0.92 &\cellcolor{blue!10} 5.46  \\  
                                        0.21 &                    0.78  \\  
                                        0.21 &                    0.78  \\
                            \end{tabular}

    \end{tabular}
    \caption{Comparison of different algorithms with ReDUCE (+Re) for CoCo (CoCo+Re), Random Forest (RF+Re), random sampling (Baseline+Re), and MIP (MIP+Re). Success rate and deterioration on the objective is denoted as S\% and Det, respectively. Average solving time and maximum solving time are denoted as Avg and Max, respectively. Int stands for the number of integer variables within a cluster. With more data the algorithms generally tend to improve performance.}
    \label{Table:Compaison_Coco_RF}
\end{table*}

\begin{figure}[!t]
		\centering
		\hspace*{-0.2cm} 
		\includegraphics[scale=0.4]{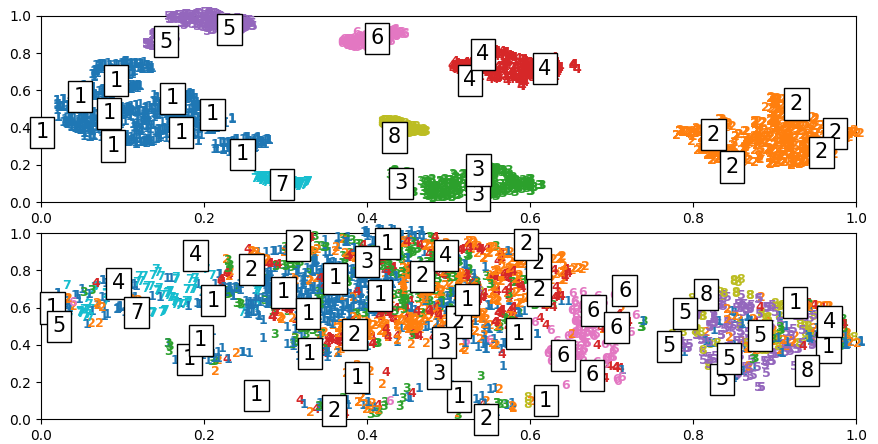}
        \hspace{2mm}
        \caption {\emph{Top}: projection onto a 2-dim manifold using t-SNE to depict clustering of a 48-dim solution space. Each color and label specify a solution clustering. Under this projection the solution appears very much structured. \emph{Bottom}: projection of the 17-dim feature space corresponding to the solution using t-SNE. The clusters have some structure over the feature space with certain labels only being found in certain regions. For visualization purposes only the top 8 clusters out of 100 are being displayed.}
		\label{Fig:clusters}
\end{figure}

We randomly sample 4,000 bookshelves and implement DBSCAN to realize 100 clusters. Fig. \ref{Fig:Figure_1} shows the first 2 dimensions of the projected solution set $X(\Theta)$ and 6 modes with distinct labels and colors. Fig.~\ref{Fig:clusters} shows the solution set packed into tighter groups compared to the same sample of features using t-distributed Stochastic Neighbor Embedding (t-SNE) \cite{van2008visualizing} as a projection of the high dimensional space to 2-dim. The upper graph in Fig.~\ref{Fig:clusters} shows the clusters in the solution space, while the lower one depicts the corresponding clusters in the feature space according to $(\Theta, \textbf{x})$ from the \textit{kick off} data. The colors in Fig.~\ref{Fig:clusters} denote the different clusters. We can tell obvious separations in the solution space. Although the clusters are more intertwined in the feature space because of the complexity of $\Theta \longrightarrow \textbf{x}$ mapping, there are still distinct regions where certain colors are more dominant. We trained a random forest (RF) classifier on the features $\Theta$. The classification accuracy reaches 97\% indicating that RF$(\{\Theta\}) \longrightarrow \{\textbf{x}\}$ is able to achieve a reasonable mapping.

\subsection{Supervised Learning}
\label{Sec:supervised_learning}
\begin{table}[!b]
    \centering
    \begin{tabular}{l|r|r|r|r}
                        & Success (\%) & Det  & Avg  Time   & Max Time   \\
    CoCo+Re         & 99.2\%       & 140  & 1.21 sec    & 13.47 sec  \\
    MIP+Re          & 100 \%       & 0    & 2.36 sec    & 48 sec     \\
    MIP                 & n/a          & n/a  & $>$ 851 sec  & n/a        \\
    MINLP               & n/a          & n/a  & $>$ 10 min  & n/a        
    \end{tabular}
    \caption{CoCo+Re (+Re denotes using ReDUCE) returned the top 50 candidate solutions over 100 clusters. MIP+Re used those same clusters to solve the problem on Gurobi. MIP also used Gurobi. MINLP ran using BONMIN as the solver. MIP and MINLP solving time for the full set of samples have exceeded reasonable limitations beyond practical purposes.
    }
\label{Tab:synthesized_results}
\end{table}

With the RF classifier trained on 100 clusters, we can quickly sample different bookshelf problems and solve the reduced MIP problem within seconds to collect more data for supervised learning. To verify the benefits with ReDUCE, we first run an experiment that observes an increase in performance with increasing amounts of data over several clusters each containing different number of integer variables (denoted as Int in Table~\ref{Table:Compaison_Coco_RF}). We pick 3 different clusters with number of integer variables 77, 66, and 46, respectively. Four different methods to solve the reduced problem within a cluster are used and tested on a fixed testing set of 500 data points. The results are shown in Table~\ref{Table:Compaison_Coco_RF}. The column titled ReDUCE shows the total number number of unique integer strategies (N) over training data (total), and the uniqueness percentage (\%) is the resulting quotient. For the column titled CoCo+Re, we train a neural network with one hidden layer of size 10,000. The input size is equal to dimension of feature space (17) and the output layer is equal to the number of unique strategies within training data (N). The neural network then samples 30 candidate strategies online according to the rank of the softmax scores of the network and solves the associated convex optimization problems. If one strategy gives a feasible solution, we terminate the process and record the solving time and optimal cost. Otherwise, infeasibility is recorded. Since the maximum number of convex problems considered is 30 to be solved online, the problem setup time is non-negligible. To avoid additional overhead we only setup the problem once and keep on modifying the integer constraints for more instances. Thus, all solving times include the problem setup time. The solving process is done on a Core i7-7800X 3.5GHz $\times$ 12 machine with Gurobi. For column RF+Re, we instead train a random forest with 150 decision trees and get the top 30 most voted strategies for candidate solutions. For column titled Baseline+Re, we simply randomly sample 30 unique strategies from the training strategies. For column MIP+Re, we solve the MIP with reduced number of integer variables, instead of using supervised learning or sampling method. For all sampling methods, we record the rate of feasibility (S\%), the deterioration (Det) of optimal cost compared to the optimal cost from MIP+Re column. That is, for the MIP+Re column, its feasibility rates are all 100\%, and its optimal deteriorations are 0. Solving times for baseline is omitted as they are significantly longer than learning based methods. All times under Avg (average solving time) and Max (maximum solving time) columns are given in seconds.

As the results show, the performances of learning based methods in general improve with more data. As ReDUCE improves the solving speed to collect larger amounts of data, we can further increase the performances with more data. The random sampling baseline (Baseline+Re) has significantly worse performance than learning methods. This column ensures that the clusters are not too small making the reduced problem too simple. For larger clusters with more integer variables (e.g. cluster 0), the learning methods demonstrate an increase in solving speed over MIP. For smaller clusters (e.g. cluster 2), MIP has faster solving speeds. There is a tradeoff between problem scale and solving methods depending on the cluster size.


For a second set of experiments, we collect data for all 100 clusters, in total 17,000, and train a neural network with the same size as above. The network is then tested on a 1,000 testing set with 100 clusters blended together. The network then samples 50 strategies. The percentage of feasibility (Success (\%)), the optimal cost deterioration (Det), the average (Avg Time) and max (Max Time) solving time are shown in Table~\ref{Tab:synthesized_results}. For comparison, we record the average and maximal solving time for MIP with ReDUCE (all clusters blended together). We also record the solving time for MIP without ReDUCE, the original formulation, and the solving time for formulation \eqref{Eqn:General_formulation} solved with BONMIN, an MINLP solver. CoCo solves faster than MIP for larger clusters which is associated with more data from the \textit{kick off} data. When sampling $D(\Theta)$, we get more samples for larger clusters. Therefore, solving time is improved when averaged. For non-reduced MIP, Table~\ref{Tab:synthesized_results} shows the average solving time over 10 samples where the solving process is interrupted if it exceeds 1,000 sec. Thus, the average solving time is at least 851 sec. It is clear that without ReDUCE, it is intractable to gather the amount of data required to train a learner of decent performance. Similar speed is seen for the MINLP solver. All the code is available at: \url{https://github.com/RoMeLaUCLA/ReDUCE.git}.


\subsection{Hardware Experiment}
\label{Sec:hardware_experiment}
We implement our optimization results on hardware with a 6 degree of freedom manipulator \cite{noh2020minimal} to insert a book onto a shelf. To automate the system, a trajectory planner is required to generate the insertion motion, which is common in practice. As the focus of this paper is on high level planning, we simplify this part by manually selecting waypoints. The hardware implementation is shown in the attached video.


\section{Conclusion, Discussion and Future Work} 
\label{Sec:conclusion}
This paper proposes ReDUCE, an algorithm that relies on unsupervised learning on relatively smaller dataset to reduce the number of integer variables and generate large amounts of strategies for supervised learners. The results demonstrated improvements in solving speed over the original MIP formulations allowing for practical implementations.


As this algorithm requires \textit{kick off} data, considering data collection is an important aspect. Beyond collecting data from simulations or hardware demonstrations, we consider automatically generating formulations for known problems and transfer knowledge to a target domain as an interesting next step. In real life, we often encounter problems with similar solutions in different contexts. Developing tools that automatically generate MIP formulations from other domains may help generate \textit{kick off} data.


It is noticed from the test results that higher feasibility rates sometime lead to lower optimality. This may be due to the optimal strategy for the testing problem not included in the training data. One future work is using generative learning approaches to combine features of different training instances to generate new strategies. Other future work include increasing the amount of data to further boost performance, increasing the amount of books to test the limit of ReDUCE, implementing parallel computing to increase online solving speed, and comparing the performance of ReDUCE with reinforcement learning algorithms.




\section*{Acknowledgements}
The authors would like to thank Abhishek Cauligi and Professor Bartolomeo Stellato for helpful discussions, and Yusuke Tanaka and Hyunwoo Nam for the assisting with hardware implementation.


{
\bibliographystyle{IEEEtran}
\bibliography{references}
}

\appendix
We explain the formulation that is used to solve the MIPs in Sec.~\ref{Sec:experiment_setup}.

For a non-convex constraint, we segment it into multiple regions and locally approximate or relax them into convex constraints. In this paper, we relax the bilinear constraints locally into convex polytopes (McCormick envelopes). Each polytope is associated with a unique combination of integer variable values, hence, mixed-integer convex constraints. Assume the number of regions used is $N$. Depending on the number of integer variables used, the formulation can generally be divided into 2 categories: 1) If the number of integer variables is $N$, we call it \textit{$N$ formulation}, e.g., the convex hull formulation \cite{belotti2011disjunctive}. 2) If the number of integer variables is $log_{2}N$, we call it \textit{$log_{2}N$ formulation}, e.g., \cite{vielma2011modeling}. \cite{vielma2011modeling} presents a \textit{$log_{2}N$ formulation} to model the special ordered sets of type 2 (sos2). However, there are several limitations of this formulation. For example, the segmented regions need to be connected to be a valid sos2 constraint, i.e., the two consecutive couple of non-zero entries can be any consecutive couple in the set. In this paper, we use MIP to model convex polytopes of arbitrary locations. This can increase the complexity for sos2 techniques as the polytopes can be disjunctive. The disjunctive constraints can be handled with convex hull formulations at a price of introducing more integer variables which may result in slower solving speeds.

In this appendix, we demonstrate an intuitive but general \textit{$log_{2}N$ formulation} to model combinations of convex polytopes at any locations which serves as the base MIP formulation for the bilinear constraints in our paper. Assume the variable $\textbf{x}$ is enforced to be within one of the $N$ convex polytopes, denoted by $\textbf{A}_{i}\textbf{x} \leq \textbf{b}_{i}$, $i=1,...,N$. We introduce $m=log_{2}N$ binary variables $z_{1}, ..., z_{m}$, $z_{i} \in \{0,1\}$. Each combination of unique values of binary variables can be assigned to a convex polytope. Let the assignment be:

\begin{equation}
    \textbf{z}=\Bar{\textbf{z}}_{i} \Rightarrow \textbf{A}_{i}\textbf{x} \leq \textbf{b}_{i}
\end{equation}

Where $\Bar{\textbf{z}}_{i} = [\Bar{z}_{i,1}, ..., \Bar{z}_{i,m}]$ are constant binary values associated with polytope $i$. Note $\Bar{z}_{i} \neq \Bar{z}_{j}$ if $i \neq j$. In other words, we require that when $\textbf{z}=\Bar{\textbf{z}}_{i}$, $\textbf{x}$ stays within the polytope $\textbf{A}_{i}\textbf{x} \leq \textbf{b}_{i}$; otherwise, the constraint is unenforced.

Denote the vertices of the polytopes by $\textbf{v}_{i,1}, ..., \textbf{v}_{i,n_{i}}$, $i=1, ..., N$, where $n_{i}$ is the number of vertices associated with polytope $i$. Each vertex $\textbf{v}_{i,j}$ is assigned a continuous non-negative variable $\lambda_{i,j} \in [0, 1]$. In general, one can run a mathematical program (e.g. \cite{avis1992pivoting}) to get vertices from the nondegenerate system of inequalities $\textbf{A}_{i}\textbf{x} \leq \textbf{b}_{i}$. As a result, the assignment becomes: 

\begin{equation}
    \textbf{z}=\Bar{\textbf{z}}_{i} \Rightarrow \ \ 
\begin{aligned}
    & \textbf{x} = \sum_{j=1}^{n_{i}} \lambda_{i,j} \textbf{v}_{i,j} \\
    & \sum_{j=1}^{n_{i}} \lambda_{i,j} = 1, \ \ \lambda_{i,j} \in [0, 1]
\end{aligned}
\label{Eqn:MICP_formulation_one}
\end{equation}

The formulation can be written as:

\begin{equation}
\begin{aligned}
    & (5.a) \quad \textbf{x} = \sum_{i=1}^{N} \sum_{j=1}^{n_{i}} \lambda_{i,j} \textbf{v}_{i,j} \\
    & (5.b) \quad  \sum_{i=1}^{N} \sum_{j=1}^{n_{i}} \lambda_{i,j} = 1, \ \ \lambda_{i,j} \in [0, 1] \\
    & (5.c) \quad  \sum_{k=1,...,N}^{k \neq i} \sum_{j=1}^{n_{k}} \lambda_{k,j} \leq \sum_{l=1}^{m} |z_{l} - \Bar{z}_{i,l}|
\end{aligned}
\label{Eqn:MICP_formulation_all}
\end{equation}

The set of constraint in \eqref{Eqn:MICP_formulation_all} enforces \eqref{Eqn:MICP_formulation_one} for all polytopes, as $\sum_{l=1}^{m} |z_{l} - \Bar{z}_{i,l}| = 0$ only when $\textbf{z}=\Bar{\textbf{z}}_{i}$, enforcing that all $\lambda$'s that are not associated with polytope $i$ to be zero. If $\textbf{z} \neq \Bar{\textbf{z}}_{i}$, $\sum_{l=1}^{m} |z_{l} - \Bar{z}_{i,l}| \geq 1$ and constraint (5.c) is looser than constraint (5.b), hence, trivial.

Note that formulation \eqref{Eqn:MICP_formulation_all} works for any convex polytope that can be written as $\textbf{A}_{i}\textbf{x}\leq \textbf{b}_{i}$. In this paper, the polytopes are McCormick envelope constraints which is a special case.

\end{document}